\pdfoutput=1
\documentclass[11pt]{article}

\usepackage[final]{acl}

\usepackage{times}
\usepackage{latexsym}

\usepackage[T1]{fontenc}
\usepackage[utf8]{inputenc}

\usepackage{microtype}

\usepackage{inconsolata}

\usepackage{graphicx}
\usepackage[utf8]{inputenc} % allow utf-8 input
\usepackage[T1]{fontenc}    % use 8-bit T1 fonts
\usepackage{hyperref}       % hyperlinks
\usepackage{url}            % simple URL typesetting
\usepackage{booktabs}       % professional-quality tables
\usepackage{amsfonts}       % blackboard math symbols
\usepackage{amsmath}
\usepackage{nicefrac}       % compact symbols for 1/2, etc.
\usepackage{microtype}      % microtypography
\usepackage{xcolor}         % colors
\usepackage[ruled,vlined,linesnumbered]{algorithm2e}
\usepackage{multirow}
\usepackage{arydshln}
\usepackage{adjustbox}
\usepackage{pifont}
\usepackage{enumitem}
\usepackage{tcolorbox}
\tcbuselibrary{listingsutf8}

\lstdefinestyle{promptstyle}{
    basicstyle=\ttfamily\small,
    breaklines=true,
    breakatwhitespace=false,
    keepspaces=false,        % 不保留空格（禁用自动缩进）
    frame=single,
    backgroundcolor=\color{gray!10},
    showstringspaces=false,
    columns=fullflexible,
    xleftmargin=0pt,
    aboveskip=1em,
    belowskip=1em
}

\newcommand{\M}{{AdaptFlow}\xspace}

\title{\M: Adaptive Workflow Optimization via Meta-Learning}

\author{
 \textbf{Runchuan Zhu\textsuperscript{1}\thanks{Equal contribution. For inquiries, please contact: zhurunchuan@stu.pku.edu.cn.}\thanks{Work is done during an internship at Microsoft}},
 \textbf{Bowen Jiang\textsuperscript{1}\footnotemark[1]},
 \textbf{Lingrui Mei\textsuperscript{2}},
 \textbf{Fangkai Yang\textsuperscript{3}\thanks{Corresponding author.}},
 \\
 \textbf{Lu Wang\textsuperscript{3}},
 \textbf{Haoxiang Gao\textsuperscript{1}},
 \textbf{Fengshuo Bai\textsuperscript{4}},
 \textbf{Pu Zhao\textsuperscript{3}},
 \\
 \textbf{Qingwei Lin\textsuperscript{3}},
 \textbf{Saravan Rajmohan\textsuperscript{3}},
 \textbf{Dongmei Zhang\textsuperscript{3}}
\\
 \textsuperscript{1}Peking University, 
 \textsuperscript{2}University of Chinese Academy of Sciences, \\
 \textsuperscript{3}Microsoft, 
 \textsuperscript{4}Shanghai Jiaotong University, \\
\\
}

\begin{document}
\maketitle
\begin{abstract}
Recent advances in large language models (LLMs) have sparked growing interest in agentic workflows, which are structured sequences of LLM invocations intended to solve complex tasks. However, existing approaches often rely on static templates or manually designed workflows, which limit adaptability to diverse tasks and hinder scalability. We propose AdaptFlow, a natural language-based meta-learning framework inspired by model-agnostic meta-learning (MAML). AdaptFlow learns a generalizable workflow initialization that enables rapid subtask-level adaptation. It employs a bi-level optimization scheme: the inner loop refines the workflow for a specific subtask using LLM-generated feedback, while the outer loop updates the shared initialization to perform well across tasks. This setup allows AdaptFlow to generalize effectively to unseen tasks by adapting the initialized workflow through language-guided modifications. Evaluated across question answering, code generation, and mathematical reasoning benchmarks, AdaptFlow consistently outperforms both manually crafted and automatically searched baselines, achieving state-of-the-art results with strong generalization across tasks and models. The source code and data are available at \url{https://github.com/microsoft/DKI_LLM/tree/AdaptFlow/AdaptFlow}.

% \lu{Recent advances in large language models (LLMs) have sparked growing interest in agentic workflows—structured sequences of LLM invocations for solving complex tasks. However, most existing approaches rely on static templates or manually designed workflows, limiting adaptability and scalability. We propose AdaptFlow, a natural language–based meta-learning framework that enables flexible and generalizable agent optimization without requiring gradient descent. AdaptFlow performs bi-level optimization entirely in natural language: the inner loop adapts to specific task instances through LLM-generated feedback, optimizing decisions at the sample level, while the outer loop aggregates these adaptations to refine the overall agent framework. This approach yields a generalizable initialization that improves performance across diverse tasks. Evaluated on question answering, code generation, and mathematical reasoning benchmarks, AdaptFlow consistently outperforms handcrafted and automatically searched baselines, achieving state-of-the-art results with strong cross-task and cross-model generalization.}

\end{abstract}

\section{Introduction}
\label{sec:1}

Recent progress in Large Language Models (LLMs)~\cite{achiam2023gpt, guo2025deepseek, Mei_2024} has led to remarkable performance across diverse tasks, including question answering~\cite{rajpurkar2016squad, yang2018hotpotqa, ding20243ds, jiang2025evaluating}, code synthesis~\cite{chen2021evaluating, nijkamp2023codegen, mei2025a1steeptesttimescaling}, and multi-turn dialogue~\cite{zhang2020dialogpt, bai2022training, 2025utilize, 2025grait}. Beyond static prediction, LLMs are increasingly being used as decision-making agents capable of dynamic reasoning and adaptive behavior~\cite{shinn2023reflexion, wei2022chain, yao2023tree}. This development has given rise to the notion of \textit{agentic workflows}, which organize LLMs into structured sequences of actions involving task decomposition, planning, tool use, execution, and self-reflection~\cite{yao2023tree, creswell2023selection}. Such workflows have demonstrated strong performance in settings that require multi-step reasoning~\cite{yao2023tree, creswell2023selection}, long-horizon planning~\cite{liu2023llmplanner, zhou2024llm}, and external tool integration~\cite{schick2023toolformer, qin2023toolllm}.

% While effective, manually designing agentic workflows is time-consuming and difficult to generalize across diverse tasks. To address this, recent work has explored automated workflow construction through prompt optimization, hyperparameter tuning, and structural search~\cite{khattab2023dspy, chen2023autoagents, li2024autoflow, liu2024dynamic, song2024adaptive, zhang2024g, wang2025scoreflow}. However, many of these approaches~\cite{liu2024dynamic, zhang2024g} rely on graph-based representations with limited support for conditional states, which restricts the expressivity of the search space and hinders their applicability to complex tasks.

While effective in controlled settings, manually designing agentic workflows is time-consuming and lacks scalability across diverse tasks. To address this, recent work has explored automated workflow construction through prompt optimization~\cite{khattab2023dspy, chen2023autoagents}, hyperparameter tuning~\cite{li2024autoflow, wang2025scoreflow}, and structural search~\cite{liu2024dynamic, song2024adaptive, zhang2024g}. However, many of these methods~\cite{liu2024dynamic, zhang2024g} represent workflows using fixed graph structures, which inherently limit the flexibility of the agentic workflow search space.

Recent frameworks such as ADAS~\cite{hu2024automated} and \textsc{Aflow}~\cite{zhang2024aflow} adopt code-based workflow representations to enable robust and flexible search. However, as noted by~\citet{wang2025scoreflow}, these methods typically generate a single static workflow for the entire task set, limiting their ability to generalize across heterogeneous datasets with diverse problem types. In addition, ADAS performs coarse-grained workflow updates, resulting in redundant context accumulation and growing complexity that hinders convergence. \textsc{Aflow} alleviates some of these issues using Monte Carlo Tree Search, but its reliance on discrete updates and early pruning can restrict the exploration of more expressive workflow candidates. These limitations underscore two challenges simultaneously:  
\textbf{\textit{C1}. How to adaptively construct effective workflows for datasets containing diverse problems?}  
\textbf{\textit{C2}. How to ensure convergent optimization in code search spaces?}

To tackle these challenges, we propose \textbf{\M}—a meta-optimization framework that incorporates principles from Model-Agnostic Meta-Learning (MAML)~\cite{finn2017model} into agentic workflow optimization. MAML learns an initialization that enables rapid adaptation to new tasks via a bi-level optimization: the \textit{inner loop} performs task-specific updates, and the \textit{outer loop} updates the initialization to generalize across tasks. Inspired by this structure, \M learns a shared workflow initialization that can quickly adapt to diverse subtasks through symbolic updates guided by LLM-generated feedback. Specifically, \M follows a \textit{bi-level optimization scheme}, where the \textit{inner loop} iteratively refines the workflow based on subtask-level feedback, while the \textit{outer loop} consolidates these refinements into a generalizable initialization. To further enhance adaptability, we perform an additional unsupervised adaptation step at test time on each target subtask, leveraging semantic descriptions derived from input prompts. This design enables both effective subtask-specific adaptation (addressing \textbf{C1}) and stable convergence in code spaces (addressing \textbf{C2}), offering a scalable solution for general-purpose workflow construction.

Our key contributions are summarized as follows:

\begin{itemize}[leftmargin=*,noitemsep,topsep=2pt]
    % \item We draw a novel analogy between neural network training and automated workflow optimization, which motivates \M design.
    
    % \item We propose \M, a natural language-based meta-learning framework that incorporates MAML to enable rapid subtask adaptation.
    \item We introduce \textbf{AdaptFlow}, a meta-learning framework that integrates the MAML paradigm with natural language supervision. AdaptFlow replaces conventional gradients with \emph{textual gradients}, which are natural language feedback generated by large language models. This mechanism enables efficient subtask-level adaptation within the programmatic code space.

    \item We design a \textit{bi-level optimization framework} tailored for code space. In the \textbf{inner loop}, workflows are iteratively refined using LLM-generated textual feedback. To ensure meaningful and stable updates, we introduce a binary continuation signal that determines whether each update leads to a non-trivial performance gain. In the \textbf{outer loop}, we aggregate subtask-level improvements into a shared initialization, further enhanced by a reflection step that revisits failure cases to improve robustness and convergence.
    
    \item Experiments on benchmarks in question answering, code generation, and mathematical reasoning show that \M outperforms both manual workflows and prior baselines, achieving state-of-the-art results with strong model-agnostic generalization.
\end{itemize}

\section{Related Work}
\label{sec:2}

\subsection{Agentic Workflow}

Agentic workflows provide a structured alternative to autonomous agents for deploying LLMs. Instead of learning through environment interaction~\cite{zhuge2023gptswarm,hong2024adaptive}, they execute static or semi-static sequences inspired by human reasoning~\cite{zhang2024aflow}, offering better interpretability and modularity.

Workflows can be general purpose, incorporating reusable patterns such as chain-of-thought prompting, self-refinement, or role decomposition~\cite{wei2022chain,shinn2023reflexion}—or domain-specific, tailored for areas such as code generation~\cite{hong2024sweagent,ridnik2024coding,zhao2024agentcoder}, data analysis~\cite{xie2024autonova,ye2024llmdp,li2024autoda}, mathematics~\cite{zhong2024math,xu2024mathematics}, and complex QA~\cite{nori2023capabilities,zhou2024reasoning}. While effective, manually designed workflows require significant human effort and lack adaptability, motivating automated optimization.

\begin{figure*}[t]
  \centering
  \includegraphics[width=1.0\linewidth]{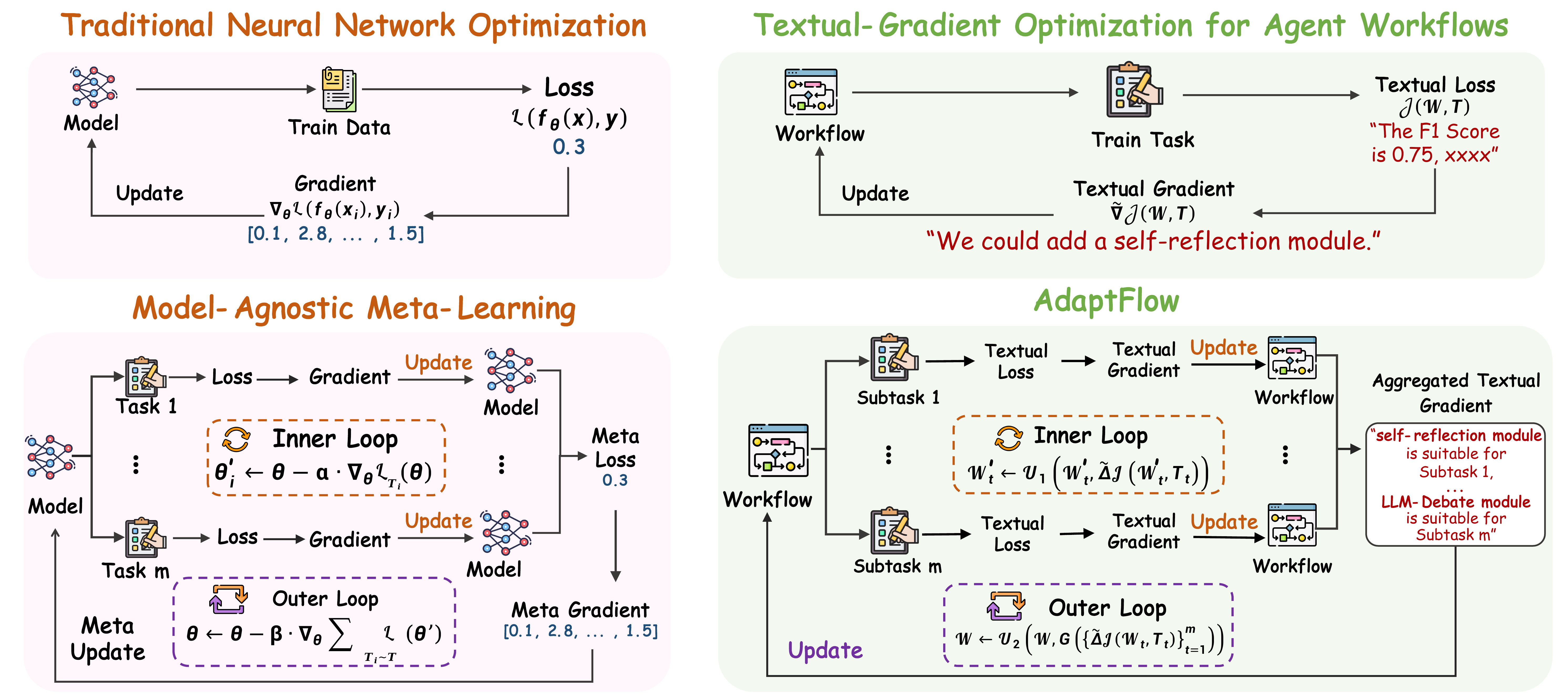}
  \caption{An analogy between Neural Network Optimization and Workflow Optimization, as well as between MAML and \M.}
  \label{fig:analogy}
\end{figure*}

\subsection{Agentic Workflow Optimization}

Recent advances~\cite{hu2024automated, zhang2024aflow, wang2025scoreflow, li2024autoflow, chen2023autoagents, song2024adaptive, hong2024sweagent} have explored automating agentic workflows to improve LLM performance. Some methods focus on optimizing prompts or parameters within fixed workflows~\cite{fernando2023promptbreeder, guo2023evoprompt, khattab2023dspy, saad2024hyperparameter}, improving reasoning without altering the execution structure. In contrast, we optimize workflow structures directly, enabling broader adaptation across tasks.

Other approaches search over code-based workflows. ADAS~\cite{hu2024automated} refines linear traces of executable code, while AFLOW~\cite{zhang2024aflow} introduces compositional abstractions with MCTS. ScoreFlow~\cite{wang2025scoreflow} frames workflow generation as supervised prediction. However, these methods often produce static workflows and lack task-level adaptability. Our method, AdaptFlow, differs by performing bi-level meta-learning: it adapts workflows via LLM feedback at the subtask level and consolidates them into a generalizable initialization, supporting fast adaptation and robust generalization.

\section{Preliminaries}
\label{sec:3}

\subsection{Problem Formulation}

The goal of automated agentic workflow optimization is to discover effective compositions of modular components—such as prompt templates, tool invocations, control logic, and reflection routines—that can guide LLMs to solve complex tasks across diverse domains.

We consider the problem of agentic workflow design, where the goal is to discover an effective workflow $\mathcal{W}$ that can solve a given task $\mathcal{T}$ drawn from a distribution. The workflow search is defined by three core components:
\begin{itemize}
    \item $\mathcal{S}$ denotes the \textit{search space}, encompassing all candidate workflows;
    \item $\mathcal{J}: \mathcal{S} \times \mathcal{T} \rightarrow \mathbb{R}$ is the \textit{objective function} that quantifies the quality or utility of a workflow $\mathcal{W} \in \mathcal{S}$ when applied to a specific task $\mathcal{T}$;
    \item $\mathcal{A}$ represents the \textit{search algorithm}, which explores $\mathcal{S}$ and generates candidate workflows guided by feedback from $\mathcal{J}$.
\end{itemize}

Given a task $\mathcal{T} \sim \mathcal{P}(\mathcal{T})$, the agent seeks to identify an optimal workflow through a task-conditioned search process:

\begin{align}
\mathcal{W} &= \mathcal{A}(\mathcal{S}, \mathcal{J}, \mathcal{T}), \\
\mathcal{W}^\star &= \mathop{\arg\max}_{\mathcal{W} \in \mathcal{S}} \; \mathbb{E}_{\mathcal{T} \sim \mathcal{P}(\mathcal{T})} \left[ \mathcal{J}(\mathcal{W}, \mathcal{T}) \right].
\end{align}

Building on prior efforts~\cite{hu2024automated}, our method defines the workflow search space directly in the code space, where candidate workflows are represented as executable programs.

\subsection{Analogy: From Supervised Learning to Agentic Workflow Optimization}

In traditional supervised learning, a model learns a parameterized function $f_\theta$ by minimizing the expected loss over labeled data $(x, y) \sim \mathcal{D}$:
\begin{align}
\theta^\star &= \mathop{\arg\min}_{\theta} \; \mathbb{E}_{(x, y) \sim \mathcal{D}} \left[ \mathcal{L}(f_\theta(x), y) \right], \\
\theta &\leftarrow \theta - \eta \cdot \frac{1}{N} \sum_{i=1}^N \nabla_\theta \mathcal{L}(f_\theta(x_i), y_i),
\end{align}
where $\eta$ is the learning rate and $\{(x_i, y_i)\}_{i=1}^N$ is a mini-batch of training examples. This process relies on differentiable loss functions and explicit ground-truth supervision, enabling gradient-based parameter updates in continuous space.

Analogously, agentic workflow optimization operates in a symbolic structure space defined over executable code (e.g.,~\cite{hu2024automated}). Given a task $\mathcal{T}$, the system executes a workflow $\mathcal{W}$ and obtains a task-level utility score from the objective function $\mathcal{J}(\mathcal{W}, \mathcal{T})$. The goal is to discover a workflow that maximizes the expected utility across a distribution of tasks:
\begin{align}
\mathcal{W} &\leftarrow \mathcal{U}_1\left(\mathcal{W}, \tilde{\nabla}\mathcal{J}(\mathcal{W}, \mathcal{T})\right).
\end{align}
Here, $\mathcal{J}(\mathcal{W}, \mathcal{T})$ denotes a natural language evaluation of a workflow’s performance on task $\mathcal{T}$, serving as a form of \textit{textual loss}. From this, the LLM generates a \textit{textual gradient} $\tilde{\nabla} \mathcal{J}$—feedback that suggests improvements, identifies failure cases, or proposes structural edits. For example, the feedback may suggest “we could add a self-reflection module” to improve performance, providing actionable guidance for workflow revision. The update operator $\mathcal{U}_1$ then applies such feedback to revise the workflow $\mathcal{W}$ in code space, enabling symbolic updates in a non-differentiable setting. This feedback-driven, interpretable optimization generalizes the notion of learning beyond standard gradient descent (Figure~\ref{fig:analogy}).

\subsection{Model-Agnostic Meta-Learning}

Model-Agnostic Meta-Learning (MAML)~\cite{finn2017model} learns a model initialization that enables rapid adaptation to new tasks using only a few gradient steps. The core idea is to train the model not just to perform well on a set of tasks, but to be easily fine-tuned for any new task drawn from the same distribution.

Given a task distribution $\mathcal{T}$, each task $\mathcal{T}_i \sim \mathcal{T}$ is associated with a loss $\mathcal{L}_{\mathcal{T}_i}(\theta)$. MAML performs a bi-level optimization:

\begin{align}
\theta_i' \leftarrow \theta - \alpha \nabla_{\theta} \mathcal{L}_{\mathcal{T}_i}(\theta), \\
\theta \leftarrow \theta - \beta \nabla_{\theta} \sum_{\mathcal{T}_i \sim \mathcal{T}} \mathcal{L}_{\mathcal{T}_i}(\theta_i').
\end{align}

In the inner loop, the model performs gradient descent on a given task to obtain adapted parameters $\theta_i'$. In the outer loop, the original initialization $\theta$ is updated using the post-adaptation losses across multiple tasks. This procedure leverages second-order gradients and enables generalization to unseen tasks with minimal fine-tuning.

\section{Methodology}
\label{sec:4}

\begin{algorithm}[t]
\SetAlgoLined
\DontPrintSemicolon
\KwIn{train tasks $\mathcal{T}_{\text{train}}$, test tasks $\mathcal{T}_{\text{test}}$, inner iterations $n_{\text{inner}}$, outer iterations $n_{\text{outer}}$}
\caption{\M Algorithm} \label{alg}
Cluster $\mathcal{T}_{\mathrm{train}}$ into $m$ subtasks $\{\mathcal{T}_1, \ldots, \mathcal{T}_m\}$; \\
Initialize global workflow $\mathcal{W}$ = $\mathcal{W}_1$ = ... = $\mathcal{W}_m$; \\
\tcp*[h]{\textbf{Outer loop}} \\
\For{$i \gets 1$ \KwTo $n_{\text{outer}}$}{
    \ForEach{$\mathcal{T}_t \in \{\mathcal{T}_1, \ldots, \mathcal{T}_m\}$}{
        Initialize $\mathcal{W}^{\prime}_t \leftarrow \mathcal{W}$; $j \leftarrow 0$; \\
        \tcp*[h]{\textbf{Inner loop}} \\
        \While{$\mathcal{J}(\mathcal{W}^{\prime}_t, \mathcal{T}_t) < \mathcal{J}(\mathcal{W}_t, \mathcal{T}_t) - \epsilon$ \textbf{and} $j < n_{\text{inner}}$}{
            Execute $\mathcal{W}^{\prime}_t$ on $\mathcal{T}_t$, obtain $\tilde{\nabla} \mathcal{J}$;\\
            $\mathcal{W}^{\prime}_t \leftarrow \mathcal{U}_1\left(\mathcal{W}^{\prime}_t, \tilde{\nabla} \mathcal{J}\right)$; \\
            $j \leftarrow j + 1$;
        }
        $\mathcal{W}_t \leftarrow \mathcal{W}^{\prime}_t$; 
    }
    $\mathcal{W} \leftarrow \mathcal{U}_2\left(\mathcal{W}, G\left(\left\{ \tilde{\nabla} \mathcal{J}(\mathcal{W}_t, \mathcal{T}_t) \right\}_{t=1}^m\right)\right)$
}
Cluster $\mathcal{T}_{\mathrm{test}}$ into $n$ subtasks $\{\mathcal{T}^{\prime}_1, \ldots, \mathcal{T}^{\prime}_n\}$; \\
\ForEach{$\mathcal{T}^{\prime}_t$}{
    $\mathcal{W}^{\prime} \leftarrow \mathcal{U}_3\left(\mathcal{W}, \mathcal{T}^{\prime}_t\right)$;\\
    Evaluate $\mathcal{W}^{*}$ on $\mathcal{T}^{\prime}_t$;
}
\end{algorithm}

\subsection{Overview}

% 基于文本 这点要着重体现

% C1: Task可能具有多样性，如果去为多样性的task高效地生成合适的workflow
% C2: Workflow 优化如何保证收敛性

% \M运用了MAML的思想，融入到Automated Design of Agentic Systems场景下，如figure 1所示，我们去优化一个可以fast adapt to 不同subtask的workflow，解决了C1. 在workflow optimaztion过程中，我们使用一种类似于MAML的双层循环结构，内循环类似于模型在做探索，类似于ADAS，但是我们通过约束保证了他的收敛性；内循环在做利用，我们实验结果表明外循环具有较好的收敛性。

\begin{figure*}[t]
  \centering
  \includegraphics[width=0.95\linewidth]{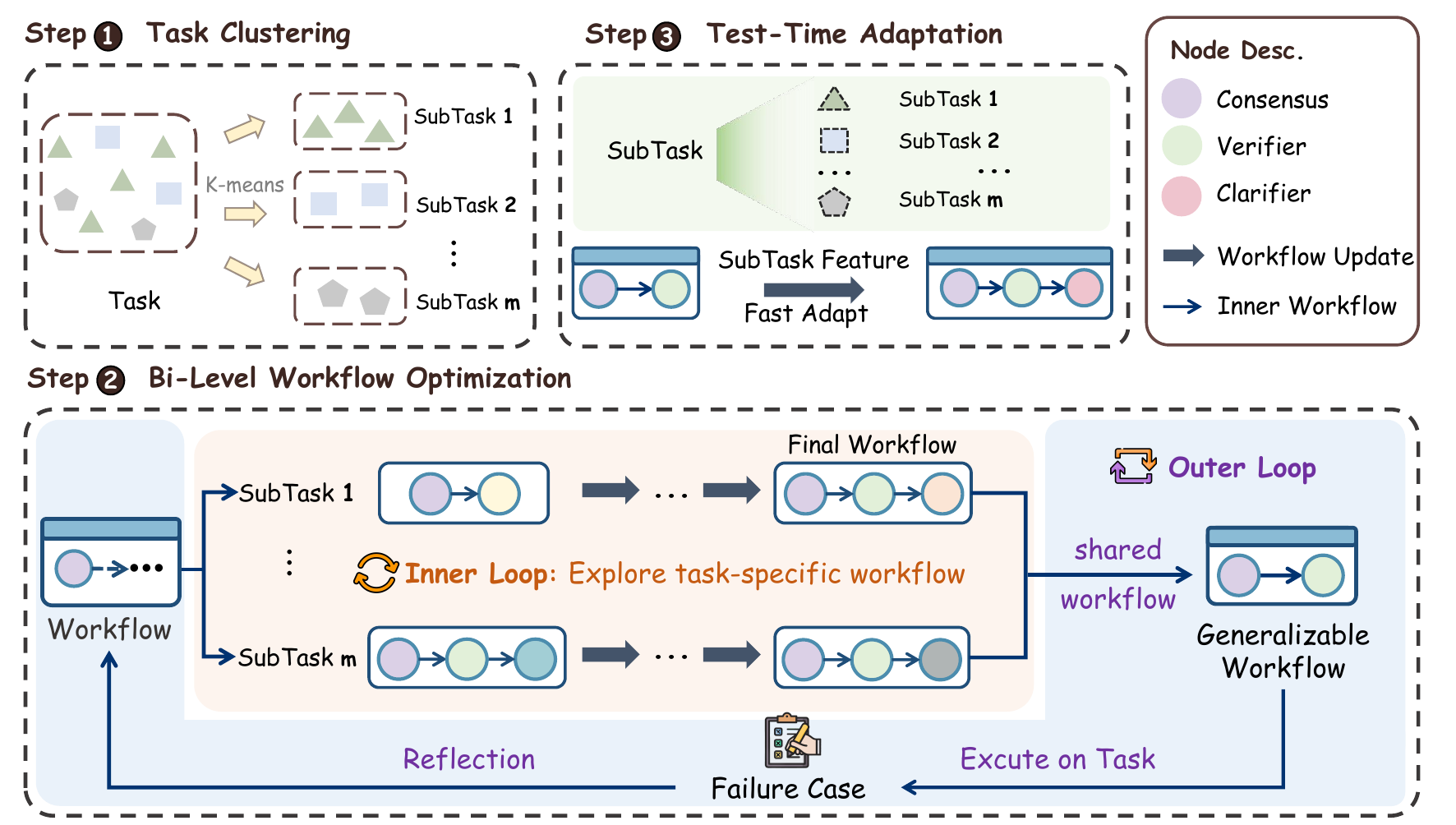}
  \caption{
Illustration of the \M framework, consisting of three stages. (1) \textbf{Task Clustering}: training tasks are grouped into semantically coherent subtasks. (2) \textbf{Bi-Level Workflow Optimization}: a bi-level optimization process is applied—inner loop explores workflow variants using LLM-generated feedback; outer loop aggregates updates into a generalizable initialization. (3) \textbf{Test-Time Adaptation}: the learned workflow is adapted to unseen tasks based on subtask-level descriptions generated from input questions. The detailed mechanism of inner and outer updates is shown in Figure~\ref{fig:analogy}.
}

  \label{fig:method}
\end{figure*}

We present \textbf{\M}, a meta-optimization framework that integrates ideas from MAML~\cite{finn2017model} into the setting of agentic workflow optimization. As illustrated in Figure~\ref{fig:method}, our method first partitions the training tasks into multiple semantically coherent subtasks. It then performs a bi-level optimization process to learn a workflow initialization that generalizes across these subtasks: the \textbf{inner loop} (lines 5–12 in Algorithm~\ref{alg}) adapts the workflow using LLM-generated feedback for each subtask, while the \textbf{outer loop} (lines 3–14 in Algorithm~\ref{alg}) aggregates these refinements into a shared initialization. At test time, we apply lightweight adaptation on unseen subtasks based on their semantic descriptions (lines 16–19  in Algorithm~\ref{alg}). By explicitly optimizing workflows at the subtask level, \M enables structural adaptation to diverse problem types, addressing challenge (\textbf{C1}). Furthermore, the hierarchical inner–outer update scheme ensures stable convergence in the discrete code space, effectively resolving challenge (\textbf{C2}). The full algorithm is provided in Algorithm~\ref{alg}, and its procedural flow is visualized in Figure~\ref{fig:method}.

% \lu{there are sevral different colors of icons, only cot and reflexion are explained. it could be better to explain all the listed icons}}

\subsection{Task Clustering}

Many tasks exhibit high internal diversity, making it difficult to optimize a single workflow across all instances. To address this, we first partition the training set $\mathcal{T}_{\text{train}}$ into $m$ semantically coherent subtasks ${\mathcal{T}_1, \ldots, \mathcal{T}_m}$ using K-Means~\cite{macqueen1967kmeans} clustering over instruction embeddings. The embeddings are obtained from the \texttt{all-MiniLM-L6-v2} model~\cite{reimers2019sentence}. This decomposition enables subtask-specific workflow optimization and promotes more stable and effective learning.

\subsection{Bi-Level Workflow Optimization}
% Our method is inspired by the MAML framework~\cite{finn2017model}, which learns an initialization that can be quickly adapted to new tasks through a bi-level optimization process. Similarly, we first partition the training tasks into subtasks, and then perform bi-level optimization: the \textbf{inner loop} explores subtask-specific workflow refinements using LLM feedback, while the \textbf{outer loop} consolidates these updates into a shared initialization that generalizes across subtasks.

\paragraph{Inner Loop (Exploration)}  
For each subtask $\mathcal{T}_t$, the workflow $\mathcal{W}^{\prime}_t$ is iteratively refined using LLM-generated textual feedback. At each step, we evaluate the current utility $\mathcal{J}(\mathcal{W}^{\prime}_t, \mathcal{T}_t)$ and apply the symbolic update:
\begin{equation}
\mathcal{W}^{\prime}_t \leftarrow \mathcal{U}_1\left(\mathcal{W}^{\prime}_t, \tilde{\nabla} \mathcal{J}(\mathcal{W}^{\prime}_t, \mathcal{T}_t)\right).
\end{equation}

To ensure stable and meaningful exploration, we define a binary continuation signal $\delta_t \in {0, 1}$ as:
\begin{equation}
\delta_t = \mathbb{I}\left[\mathcal{J}(\mathcal{W}_t, \mathcal{T}_t) - \mathcal{J}(\mathcal{W}^{\prime}_t, \mathcal{T}_t) > \epsilon\right],
\end{equation}
where $\mathcal{W}_t$ denotes the best workflow found so far. Here, $\mathcal{J}$ evaluates a workflow’s performance on task $\mathcal{T}_t$ via textual assessment, serving as a form of task-level \textit{textual loss}. Based on this evaluation, the LLM generates a \textit{textual gradient} $\tilde{\nabla} \mathcal{J}$ that reflects potential improvements or corrections. The update operator $\mathcal{U}_1$ applies this feedback to revise the workflow $\mathcal{W}'_t$ in the code space. The inner loop continues only if $\delta_t = 1$, indicating that the update yields a non-trivial gain. This continuation signal acts as a local convergence criterion, mitigating instability from long-context accumulation and ensuring effective symbolic refinement. The prompt design for $\mathcal{U}_1$ is detailed in Section~\ref{subsubsec:inner_prompt}.

\paragraph{Outer Loop (Exploitation)}
After inner-loop optimization across all subtasks, we aggregate the resulting feedback to update the global workflow. Each $\tilde{\nabla} \mathcal{J}(\mathcal{W}_t, \mathcal{T}_t)$ denotes a textual gradient—natural language feedback from the LLM that suggests workflow improvements based on subtask performance. The aggregation function $G$ merges these gradients into a unified signal, which is then applied via the update operator $\mathcal{U}_2$:
\begin{equation}
\mathcal{W} \leftarrow \mathcal{U}_2\left(\mathcal{W}, G\left(\left\{\tilde{\nabla} \mathcal{J}(\mathcal{W}_t, \mathcal{T}_t)\right\}_{t=1}^m\right)\right).
\end{equation}
This meta-level update integrates subtask-specific insights into a generalizable workflow by aggregating the textual gradients from the best-performing workflows of each subtask and applying them to revise the global workflow.

To further improve robustness, we apply a \textbf{reflection} step after the update. The updated workflow is re-executed on each subtask to identify remaining failure cases. The agent then generates refinement suggestions, which are used to perform a secondary symbolic update. This reflection-enhanced outer loop helps address blind spots and improve generalization.

\subsection{Test-Time Adaptation}

% To evaluate generalization, we apply the learned initialization $\mathcal{W}$ to a set of unseen test tasks $\mathcal{T}_{\text{test}}$. Following the same procedure as in training, we partition $\mathcal{T}_{\text{test}}$ into $n$ subtasks $\{\mathcal{T}_1', \ldots, \mathcal{T}_n'\}$ using instruction-level clustering.

% For each subtask $\mathcal{T}_t'$, we randomly sample a subset $\tilde{\mathcal{T}}_t' \subset \mathcal{T}_t'$ and prompt a language model to generate a high-level description based on the sampled examples. This yields a subtask-level representation $\mathcal{F}(\tilde{\mathcal{T}}_t')$, which is then used to adapt the global workflow via a symbolic update operator:
% \begin{equation}
% \mathcal{W}^{*} \leftarrow \mathcal{U}_3\left(\mathcal{W}, \mathcal{F}(\tilde{\mathcal{T}}_t')\right).
% \end{equation}

% The resulting specialized workflow $\mathcal{W}^{*}$ is then evaluated on the full subtask $\mathcal{T}_t'$, enabling adaptation to previously unseen task distributions.
To evaluate generalization, we apply the learned initialization $\mathcal{W}$ to a set of unseen test tasks $\mathcal{T}{\text{test}}$. Following the same procedure as in training, we partition $\mathcal{T}{\text{test}}$ into $n$ subtasks ${\mathcal{T}_1', \ldots, \mathcal{T}_n'}$ using instruction-level clustering.

For each subtask $\mathcal{T}_t'$, we randomly sample a subset $\tilde{\mathcal{T}}_t' \subset \mathcal{T}_t'$ and prompt a language model to generate a high-level description $\mathcal{F}(\tilde{\mathcal{T}}_t')$ based solely on the input questions from the sampled tasks—without access to answers or solutions. This representation captures the subtask’s semantic intent and guides adaptation.

We then apply the update operator $\mathcal{U}_3$ to specialize the global workflow based on this subtask description:
\begin{equation}
\mathcal{W} \leftarrow \mathcal{U}_3\left(\mathcal{W}, \mathcal{F}(\tilde{\mathcal{T}}_t')\right).
\end{equation}
Here, $\mathcal{U}_3$ performs a fast adaptation of the workflow by leveraging the semantic intent of the subtask, which is derived from input prompts. It uses natural language cues to specialize the global workflow for the target subtask. The prompt design for $\mathcal{U}_3$ is detailed in Section~\ref{subsubsec:adapt_prompt}. The resulting adapted workflow $\mathcal{W}$ is then evaluated on the full subtask $\mathcal{T}_t'$, enabling effective generalization to previously unseen task distributions. A concrete example of this process is illustrated in Section~\ref{chap:6.5}.

\section{Experiment Setup}
\label{sec:5}

% TODO：补充下MATH和Olypiad的分类
\paragraph{Datasets}  
We evaluate our method on eight public datasets across three domains: question answering, code generation, and mathematical reasoning. For \textsc{HumanEval}~\cite{chen2021evaluating} and \textsc{MBPP}~\cite{austin2021program}, we use the full datasets. Following \textsc{AFLOW}~\cite{zhang2024aflow}, we sample 1,319 examples from the \textsc{GSM8K} test split~\cite{cobbe2021gsm8k}. For \textsc{MATH}~\cite{hendrycks2021measuring}, we follow~\cite{hong2024data} and select level-5 problems from four categories: Combinatorics and Probability, Number Theory, Pre-algebra, and Pre-calculus. We also include two advanced math benchmarks: \textsc{AIME}~\cite{openai2023aime} and \textsc{OlympiadBench}~\cite{zhu2024olympiadbench}. For \textsc{DROP}~\cite{dua2019drop} and \textsc{HotpotQA}~\cite{yang2018hotpotqa}, we follow prior work~\cite{shinn2023reflexion, zhang2024aflow, wang2025scoreflow} and randomly sample 1{,}000 instances each. All datasets are split into validation and test sets with a 1:4 ratio. See Table~\ref{tab:data_stats} for full statistics.

\paragraph{Baselines}  
We compare our method against two categories of baselines: manually designed workflows and automatically optimized workflows for LLMs. \textbf{Manual Workflows} include widely used prompting strategies and agent-based methods: Vanilla prompting, Chain-of-Thought (CoT)~\cite{wei2022chain}, Reflexion~\cite{shinn2023reflexion}, LLM Debate~\cite{du2023symmetric}, Step-back Abstraction~\cite{zhou2023least}, Quality-Diversity (QD)~\cite{wang2023large}, and Dynamic Role Assignment~\cite{qian2023roleplay}. These approaches are constructed using fixed templates or heuristics without task-specific adaptation. \textbf{Automatically Optimized Workflows} are derived through workflow optimization or search. We include ADAS~\cite{hu2024automated} and \textsc{AFLOW}~\cite{zhang2024aflow}, which learn or search for agentic workflow structures in a data-driven manner to improve LLM performance across tasks.

\paragraph{Implementation Details}
We use a decoupled architecture separating optimization and execution. GPT-4.1~\cite{openai2024gpt41} serves as the optimizer, while executors include DeepSeekV2.5~\cite{deepseek2024v2.5}, GPT-4o-mini~\cite{openai2024gpt4omini}, Claude-3.5-Sonnet~\cite{anthropic2024claude35}, and GPT-4o~\cite{openai2024gpt4o}. All models are accessed via public APIs with a fixed temperature of 0.5. The outer loop runs for 3 iterations, and the inner loop allows up to 6 updates per subtask.

\paragraph{Metrics}
We adopt task-specific evaluation metrics tailored to each dataset category. For mathematics benchmarks, including \textsc{GSM8K}, \textsc{MATH}, \textsc{AIME}, and \textsc{OlympiadBench}, we use the \textbf{Solve Rate}—the proportion of correctly solved problems—as the primary metric. For code generation tasks (\textsc{HumanEval} and \textsc{MBPP}), we report \textbf{pass@1}, following the evaluation protocol of Chen et al.~\cite{chen2021evaluating}, which measures the correctness of the top-1 generated solution. For question-answering datasets such as \textsc{HotpotQA} and \textsc{DROP}, we adopt the \textbf{F1 Score} to evaluate the overlap between predicted and ground-truth answers.

% GPT Swarm跑一下
% ADAS上下文冗余

\section{Results and Analysis}

\subsection{Main Results}

\begin{table*}[]
\centering
\small
\begin{adjustbox}{max width=\textwidth}
\begin{tabular}{c|cc|cc|cccc|c}
\hline
\multirow{2}{*}{Method} 
  & \multicolumn{2}{c|}{QA} 
  & \multicolumn{2}{c|}{Coding} 
  & \multicolumn{4}{c|}{MATH} 
  & \multirow{2}{*}{Average} \\
& \textsc{HotpotQA} & \textsc{DROP}
& \textsc{HumanEval} & \textsc{MBPP}
& \textsc{GSM8K} & \textsc{MATH} & \textsc{AIME} & \textsc{Olympiad} 
& \\ \hline

Vanilla              & 70.7 & 79.6 & 87.0 & 71.8 & 92.7 & 48.2 & 12.4 & 25.0 & 60.9 \\
COT                  & 69.0 & 78.8 & 90.8 & 72.5 & 91.3 & 49.9 & 10.1 & 26.4 & 61.1 \\
Reflexion            & 68.3 & 79.5 & 86.3 & 72.4 & 92.4 & 49.3 & 10.5 & 25.9 & 60.6 \\
LLM Debate           & 68.5 & 79.3 & 90.8 & 73.3 & \underline{93.8} & 52.7 & 13.7 & \underline{29.8} & 62.7 \\
Step-back Abstraction& 67.9 & 79.4 & 87.8 & 71.9 & 90.0 & 47.9 & 4.8  & 19.3 & 58.6 \\
Quality Diversity    & 69.3 & 79.7 & 88.5 & 72.5 & 92.3 & 50.5 & 9.4  & 28.8 & 61.4 \\
Dynamic Assignment   & 67.9 & 76.8 & 89.3 & 71.5 & 89.2 & 50.7 & 12.7 & 27.6 & 60.7 \\
\cdashline{1-10}
ADAS                 & 64.5 & 76.6 & 82.4 & 53.4 & 90.8 & 35.4 & 10.4 & 21.2 & 54.3 \\
AFlow                & \underline{73.5} & \underline{80.6} & \textbf{94.7} & \underline{83.4} & 93.5 & \underline{56.2} & \underline{17.4} & 28.5 & \underline{65.6} \\
\cdashline{1-10}
Ours                 & \textbf{73.8} & \textbf{82.4} & \textbf{94.7} & \textbf{84.0} & \textbf{94.6} & \textbf{61.5} & \textbf{22.6} & \textbf{34.4} & \textbf{68.5} \\ \hline
\end{tabular}
\end{adjustbox}
\caption{Performance comparison across three domains: question answering, code generation, and mathematics. Best results are shown in \textbf{bold}, and second-best results are \underline{underlined}. In our method, GPT-4.1 is used for workflow refinement, while GPT-4o-mini-0718 is responsible for workflow execution.}
\label{tab:main_exp}
\end{table*}

As shown in Table~\ref{tab:main_exp}, our method delivers consistently strong performance across three distinct domains—question answering, code generation, and mathematics—achieving the highest overall average score of \textbf{68.5}. This suggests that our unified framework generalizes well to tasks with varying structures and reasoning demands. In particular, the substantial gains on mathematics benchmarks demonstrate the framework’s strength in handling complex symbolic and multi-step reasoning.

These results highlight the advantage of learning workflows in a task-adaptive and optimization-aware manner. Compared to existing baselines, including both manually designed strategies and automatically optimized methods, our approach achieves more balanced improvements across domains, underscoring its robustness and scalability. The consistent lead over ADAS~\cite{hu2024automated} and \textsc{Aflow}~\cite{li2024autoflow}, which operate in a similar code-based search space, further supports the effectiveness of meta-level adaptation in building generalizable agentic workflows.

\subsection{Ablation Study}

\paragraph{Ablation on Reflection}
To evaluate the impact of the reflection module in the outer loop, we conduct an ablation study on the \textsc{MATH} dataset. We use GPT-4.1 for workflow updates and GPT-4o-mini-0718 for workflow execution. In the ablated setting, denoted as \texttt{w/o} reflection, we remove the reflection step where the model samples and revises failed cases after the initial outer-loop update. As shown in Table~\ref{tab:reflection_comparison}, incorporating reflection consistently leads to better performance across iterations, with a final accuracy of 61.5 compared to 60.2 without reflection. This highlights the importance of targeted self-correction in enhancing workflow robustness and adaptability.

\begin{table}[h]
\centering
\begin{tabular}{cccc}
\toprule
\textbf{Outer Loop Iteration} & \textbf{1} & \textbf{2} & \textbf{3} \\
\midrule
\texttt{w/o} reflection & 56.7 & 58.2 & 60.2 \\
\textbf{ours} & \textbf{57.2} & \textbf{58.6} & \textbf{61.5} \\
\bottomrule
\end{tabular}
\caption{Performance comparison across iterations on the \textsc{MATH} dataset. \texttt{w/o} reflection denotes the setting without the reflection component, while \textbf{ours} includes it.}
\label{tab:reflection_comparison}
\end{table}

\paragraph{Ablation on Test-Time Adaptation}

\begin{table}[h]
\centering
\begin{tabular}{ccc}
\toprule
\textbf{Subtask} & \textbf{w/o adaptation} & \textbf{ours} \\
\midrule
\textbf{PreA} & 73.1 & \textbf{76.4} \\
\textbf{PreC} & 20.8 & \textbf{21.4} \\
\textbf{C\&P} & 61.9 & \textbf{63.1} \\
\textbf{NT} & 68.3 & \textbf{73.9} \\
\cdashline{1-3}
\textbf{Overall} & 58.0 & \textbf{61.5} \\
\bottomrule
\end{tabular}
\caption{Ablation results on math subtasks with and without test-time adaptation. \texttt{w/o} adaptation disables test-time adaptation.
Subtask abbreviations: \textbf{PreC} = Precalculus, \textbf{PreA} = Prealgebra, \textbf{NT} = Number Theory, \textbf{C\&P} = Counting \& Probability. }
\label{tab:reflection_adaptation}
\end{table}

To assess the effectiveness of our test-time adaptation strategy, we conduct an ablation study on four mathematical reasoning subtasks: Prealgebra, Precalculus, Counting \& Probability, and Number Theory. As shown in Table~\ref{tab:reflection_adaptation}, removing the adaptation module results in a consistent drop in performance across all subtasks. Notably, the largest improvement is observed in Number Theory, where accuracy increases from 68.3 to 73.9, suggesting that adaptation plays a crucial role in handling complex symbolic reasoning. The overall average accuracy improves by 3.5 points, confirming that test-time refinement enhances the generalization of the global workflow to previously unseen problems.

\subsection{Convergence Analysis}

\begin{figure}[t]
  \centering
  \includegraphics[width=1.0\linewidth]{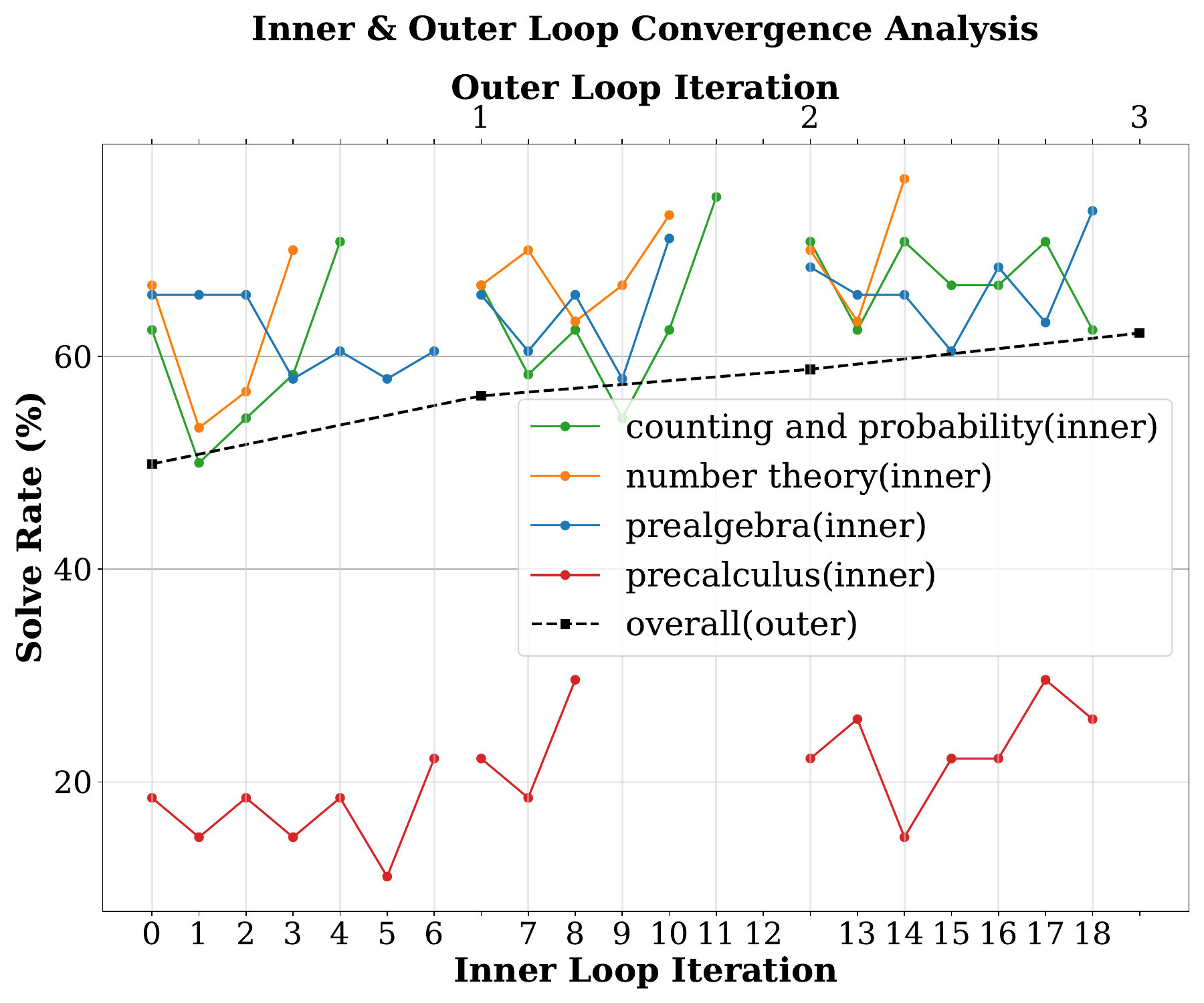}
  \caption{Convergence behavior of the inner and outer optimization loops on the \textsc{MATH} dataset. The inner loop (solid lines) shows fluctuations in solve rate across iterations for each subtask, with a maximum of 6 iterations per subtask, while the outer loop (dashed line) steadily improves overall performance by aggregating the best workflows per subtask.}
  \label{fig:convergence}
\end{figure}

We analyze the convergence behavior of both inner and outer loops on the \textsc{MATH} dataset, as shown in Figure~\ref{fig:convergence}. The inner loop exhibits noticeable fluctuations due to the accumulation of long-context dependencies and the large workflow search space, a challenge also observed in ADAS~\cite{hu2024automated}. Despite this, our constrained update mechanism helps maintain reasonable performance at each step. In contrast, the outer loop shows steady improvement, as it only aggregates the best-performing workflows from each subtask, leading to more stable and reliable updates at the meta level. These results demonstrate that our method effectively ensures convergence throughout the optimization process, addressing the core challenge of \textbf{C2}.

\subsection{Model Agnostic Analysis}
\begin{table*}[h]
\centering
\small
\begin{adjustbox}{max width=\textwidth}
\begin{tabular}{c|cccccccc}
\hline
\multirow{2}{*}{Model} & \multicolumn{8}{c}{Method}                                                                                   \\
                       & Vanilla & COT  & Reflexion & LLM debate & Step-back Abstraction & Quality Diversity & Role Assignment & Ours \\ \hline
GPT-4o-mini            & 48.2    & 49.9 & 49.3      & 52.7       & 47.9                  & 50.5              & 50.7            & \textbf{61.5} \\
GPT-4o                 & 53.8    & 53.7 & 54.2      & 55.1       & 53.3                  & 56.6              & 53.3            & \textbf{63.6} \\
claude-3-5-sonnet      & 20.4    & 22.6 & 22.6      & 23.8       & 20.7                  & 21.4              & 20.1            & \textbf{27.8} \\
DeepSeek-V2.5          & 52.6    & 52.0 & 53.3      & 54.1       & 52.8                  & 55.1              & 53.5            & \textbf{61.1} \\ \hline
\end{tabular}
\end{adjustbox}
\caption{Model-agnostic performance comparison across various workflow optimization methods on the \textsc{MATH} dataset. \textbf{Ours} consistently achieves the highest accuracy across all LLM backbones.}
\label{tab:model_agnostic}
\end{table*}

To assess generality, we evaluate our method on the \textsc{MATH} dataset using four LLMs: GPT-4o-mini, GPT-4o, Claude-3.5-Sonnet, and DeepSeek-V2.5. As shown in Table~\ref{tab:model_agnostic}, our method consistently achieves the best performance, demonstrating strong robustness and generalization.

While absolute performance varies across LLMs, our method consistently outperforms all baselines. The lower accuracy of Claude-3.5-Sonnet may stem from its weaker handling of structured outputs like JSON, which are central to our answer extraction pipeline. Nonetheless, our approach remains effective across model families without requiring model-specific customization.

\subsection{Case Study}
\label{chap:6.5}
\begin{table}[ht]
\centering
\renewcommand{\arraystretch}{1.2}
\small
\begin{adjustbox}{max width=\textwidth}
\begin{tabular}{c|ccccc}
\hline
\textbf{Module} & \textbf{All} & \textbf{PreC} & \textbf{PreA} & \textbf{NT} & \textbf{C\&P} \\
\hline
\textbf{DA} & \centering \ding{51} & \ding{51} & \ding{51} & \ding{51} & \ding{51} \\
\textbf{AE} & \ding{51} & \ding{51} & \ding{51} & \ding{51} & \ding{51} \\
\textbf{CS} & \ding{51} & \ding{51} & \ding{51} & \ding{51} & \ding{51} \\
\textbf{VF} & \ding{51} & \ding{51} & \ding{55} & \ding{55} & \ding{55} \\
\textbf{CL} & \ding{51} & \ding{55} & \ding{55} & \ding{55} & \ding{55} \\
\textbf{SY} & \ding{51} & \ding{51} & \ding{51} & \ding{51} & \ding{51} \\
\textbf{VT} & \ding{55} & \ding{55} & \ding{55} & \ding{51} & \ding{55} \\
\textbf{AD} & \ding{55} & \ding{55} & \ding{51} & \ding{55} & \ding{55} \\
\hline
\end{tabular}
\end{adjustbox}
\caption{
Module usage across subtasks on the \textsc{MATH} dataset. Each column represents a workflow configuration: 
\textbf{All} denotes the final workflow obtained after the third round of outer-loop optimization, while the others reflect the best inner-loop workflows before aggregation. 
Subtask abbreviations: \textbf{PreC} = Precalculus, \textbf{PreA} = Prealgebra, \textbf{NT} = Number Theory, \textbf{C\&P} = Counting \& Probability. 
Module abbreviations: \textbf{DA} = Diverse Agents, \textbf{AE} = Answer Extraction, \textbf{CS} = Consensus, \textbf{VF} = Verifier, \textbf{CL} = Clarifier, \textbf{SY} = Synthesis, \textbf{VT} = Value Tracker, \textbf{AD} = Approximation Detector. 
\ding{51} indicates module is used; \ding{55} indicates not used.
}
\label{tab:case_study}
\end{table}

We present a case study on the MATH dataset by comparing workflows before and after the third outer-loop iteration. Specifically, we select the best-performing workflows for each subtask prior to the final aggregation, and denote the post-aggregation unified workflow as \textbf{All}. This case study illustrates how the outer loop consolidates subtask-specific refinements into a generalizable workflow (Table~\ref{tab:case_study}). The \textbf{All} column represents the workflow obtained after the third outer-loop update, while the other columns correspond to the best inner-loop workflows before this update.

\paragraph{Shared Front-End.}
All workflows include three core modules: \textbf{DA} (Diverse Agents), \textbf{AE} (Answer Extraction), and \textbf{CS} (Consensus). These ensure solution diversity, consistent answer formats, and stable outputs, forming a robust foundation applicable across domains.

\paragraph{Task-Specific Modules.}
Additional modules are selectively introduced based on subtask characteristics. For example, \textbf{AD} (Approximation Detector) in Prealgebra handles rounding mismatches, while \textbf{VT} (Value Tracker) in Number Theory tracks intermediate values in multi-step reasoning.

This modular design supports both generalization and specialization, enabling high performance across diverse mathematical tasks.

\section{Conclusion}

We introduced \M, a bi-level meta-optimization framework that learns adaptable agentic workflows via LLM-guided symbolic feedback. Across eight benchmarks, \M outperforms both manual and automated baselines, with components like reflection and test-time adaptation enhancing robustness. Overall, it offers a scalable, model-agnostic solution for automating workflow design.

\section*{Limitations}

While \M achieves strong generalization, it has two primary limitations. First, the quality of symbolic updates depends on LLM-generated textual feedback, which can be vague or insufficiently detailed for complex failure cases. More structured or fine-grained feedback could improve update precision. Second, the optimization process requires repeated LLM queries, leading to non-trivial computational costs. Reducing query overhead through more efficient adaptation strategies is an important direction for future work.

% Bibliography entries for the entire Anthology, followed by custom entries
%\bibliography{anthology,custom}
% Custom bibliography entries only
\bibliography{main}

\appendix

\section{Appendix}
\label{sec:appendix}

\subsection{Dataset Details}

Our experiments span eight public benchmarks across three major domains: question answering, code generation, and mathematical reasoning. Table~\ref{tab:data_stats} summarizes the dataset statistics, including the number of validation/test instances and the number of subtasks for each dataset. Each subtask represents a semantically or structurally coherent group of problems, enabling more focused workflow specialization during meta-optimization.

For question answering, we use subsets of \textsc{HotpotQA} and \textsc{DROP}, each containing 1,000 examples in total, with a 1:4 split for validation and testing. The examples are clustered into six subtasks based on instruction similarity. Similarly, in the coding domain, \textsc{HumanEval} and \textsc{MBPP} are divided into three and four subtasks, respectively, reflecting different code generation patterns.

In the mathematics domain, the datasets exhibit more diverse task structures. For \textsc{GSM8K} and \textsc{AIME}, we apply instruction-level clustering to derive six distinct subtasks per dataset, capturing variations in reasoning complexity and problem format.

Notably, two datasets—\textsc{MATH} and \textsc{OlympiadBench}—come with predefined topic categories, and thus do not undergo clustering. The \textsc{MATH} dataset contains high school-level math problems and is partitioned into four canonical categories: \textbf{Prealgebra}, \textbf{Precalculus}, \textbf{Number Theory}, and \textbf{Counting \& Probability}, following the protocol introduced by \citet{hendrycks2021measuring}. These categories capture distinct types of mathematical reasoning, from basic arithmetic to combinatorial logic.

Likewise, \textsc{OlympiadBench} is sourced from competitive mathematics exams and is naturally divided into four topics: \textbf{Algebra}, \textbf{Combinatorics}, \textbf{Geometry}, and \textbf{Number Theory}, as defined in the original benchmark by \citet{zhu2024olympiadbench}. These topics correspond to challenging mathematical reasoning tasks requiring manipulation, multi-step derivation, and rigorous abstraction.

Overall, our dataset setup provides a rich and heterogeneous landscape for evaluating workflow generalization, supporting both cluster-derived and taxonomy-preserving subtask definitions across domains.

\begin{table*}[t]
\centering
\small  % 缩小字体
\renewcommand{\arraystretch}{1.2}
\begin{tabular}{l|cc|cc|cccc}
\hline
\multirow{2}{*}{} & \multicolumn{2}{c|}{QA} & \multicolumn{2}{c|}{Coding} & \multicolumn{4}{c}{MATH} \\
& \textsc{HotpotQA} & \textsc{DROP} & \textsc{HumanEval} & \textsc{MBPP} & \textsc{GSM8K} & \textsc{MATH} & \textsc{AIME} & \textsc{OlympiadBench} \\
\hline
Validation Size     & 200 & 200 & 33 & 86 & 264 & 119 & 91 & 51 \\
Val. Subtasks       & 6   & 6   & 3  & 4  & 6   & 4   & 6  & 4  \\
Test Size           & 800 & 800 & 131 & 341 & 1055 & 486 & 373 & 212 \\
Test Subtasks       & 6   & 6   & 3   & 4   & 6   & 4   & 6   & 4 \\
\hline
\end{tabular}
\caption{Dataset statistics for each domain and subtask. Validation/test sizes represent the number of instances used for evaluation, and subtask numbers denote the total distinct subtasks grouped under each benchmark.}
\label{tab:data_stats}
\end{table*}

\subsection{Analogy Explanation}

Figure~\ref{fig:method} visualizes the analogy between neural network optimization and workflow optimization, which forms the conceptual foundation for our method. Here, we detail the core correspondences both at the structure level (parameters, updates, gradients) and at the algorithmic level (meta-learning procedure).

\paragraph{Structure-Level Analogy.}
In traditional supervised learning, model training involves continuous optimization of parameters $\theta$ using gradients $\nabla_\theta L$ derived from a differentiable loss. In contrast, our workflow optimization operates in a discrete, space, where the workflow $\mathcal{W}$ is updated through textual feedback generated by LLMs. The following table presents the one-to-one mapping:

\begin{table*}[h]
\centering
\small
\renewcommand{\arraystretch}{1.3}
\begin{tabular}{l|l}
\hline
\textbf{Neural Network Optimization} & \textbf{Workflow Optimization (AdaptFlow)} \\
\hline
Model parameters $\theta$ & Workflow structure $\mathcal{W}$ \\
Loss function $L(f_\theta(x), y)$ & Utility function $J(\mathcal{W}, T)$ \\
Gradient $\nabla_\theta L$ & Textual gradient $\widetilde{\nabla} J$ (LLM feedback) \\
Gradient descent update $\theta \leftarrow \theta - \eta \nabla_\theta L$ & Symbolic update $\mathcal{W}' \leftarrow \mathcal{U}_1(\mathcal{W}, \widetilde{\nabla} J)$ \\
Batch of examples $\{(x_i, y_i)\}$ & Batch of tasks or subtask data $\mathcal{T}_t$ \\
\hline
\end{tabular}
\caption{Structure-level analogy between differentiable model optimization and discrete workflow optimization.}
\label{tab:analogy_structure}
\end{table*}

\paragraph{Meta-Learning Analogy: MAML vs. \M.}
At the algorithmic level, AdaptFlow is inspired by Model-Agnostic Meta-Learning (MAML), but adapted to the setting. While MAML learns a parameter initialization $\theta$ that can rapidly adapt via gradient updates, AdaptFlow learns a generalizable workflow $\mathcal{W}$ that adapts via LLM-generated updates. The table below compares the two approaches step-by-step:

\begin{table*}[h]
\centering
\small
\renewcommand{\arraystretch}{1.3}
\begin{tabular}{l|l}
\hline
\textbf{MAML (Finn et al., 2017)} & \textbf{AdaptFlow (Ours)} \\
\hline
Model initialization $\theta$ & Workflow initialization $\mathcal{W}$ \\
Task-specific adaptation via $\theta' \leftarrow \theta - \alpha \nabla_\theta L_{T}$ & Subtask-specific refinement via $\mathcal{W}' \leftarrow \mathcal{U}_1(\mathcal{W}, \widetilde{\nabla} J)$ \\
Compute outer gradient from $\theta'$ & Aggregate textual feedback from refined workflows $\{\widetilde{\nabla} J_t\}$ \\
Outer update: $\theta \leftarrow \theta - \beta \nabla_\theta \sum L_{T_i}(\theta'_i)$ & Meta update: $\mathcal{W} \leftarrow \mathcal{U}_2(\mathcal{W}, G(\{\widetilde{\nabla} J_t\}))$ \\
Adaptation via differentiable gradient & Adaptation via textual feedback \\
Few-shot generalization to new tasks & Test-time adaptation via $\mathcal{W}^* \leftarrow \mathcal{U}_3(\mathcal{W}, \mathcal{F}(T'_t))$ \\
\hline
\end{tabular}
\caption{Algorithm-level comparison between MAML and AdaptFlow.}
\label{tab:analogy_maml}
\end{table*}

Together, these analogies highlight how AdaptFlow generalizes the principles of meta-learning to the domain of agentic workflow optimization in spaces.

\subsection{Prompt Templates}
\subsubsection{Inner Loop Workflow Optimization Prompt}
\label{subsubsec:inner_prompt}

\UseRawInputEncoding
\begin{lstlisting}[style=promptstyle]
# Overview
You are an expert machine learning researcher testing various agentic systems. Your objective is to design building blocks such as prompts and control flows within these systems to solve complex tasks. Your aim is to design an optimal agent performing well on the MATH dataset, which evaluates mathematical problem-solving abilities across various mathematical domains including algebra, counting and probability, geometry, intermediate algebra, number theory, prealgebra and precalculus.

## An example question from MATH:

**instruction (Not Given)**: Solve the following problem and provide a detailed solution. Present the final answer using the \boxed{} format.

**question**: question

**solution (Not Given)**: solution

# Discovered architecture archive
Here is the archive of the discovered architectures:

[ARCHIVE]

The fitness value is defined as the accuracy on a validation question set. Your goal is to maximize this fitness. You should use your own judgment to decide whether to optimize on the latest architecture, as its performance may not necessarily be better.

# Output Instruction and Example:
The first key should be ("thought"), and it should capture your thought process for designing the next function. In the "thought" section, first reason about what should be the next interesting agent to try, then describe your reasoning and the overall concept behind the agent design, and finally detail the implementation steps.
The second key ("name") corresponds to the name of your next agent architecture. 
Finally, the last key ("code") corresponds to the exact “forward()” function in Python code that you would like to try. You must write a COMPLETE CODE in "code": Your code will be part of the entire project, so please implement complete, reliable, reusable code snippets.

Here is an example of the output format for the next agent architecture:

[EXAMPLE]

You must use the exact function interface used above. You need to specify the instruction, input information, and the required output fields for various LLM agents to do their specific part of the architecture. Also, it could be helpful to set the LLM’s role and temperature to further control the LLM’s response. Note that the LLMAgentBase() will automatically parse the output and return a list of “Infos”. You can get the content by Infos.content. DO NOT FORGET the taskInfo input to LLM if you think it is needed, otherwise LLM will not know about the task.

# Your task
You are deeply familiar with LLM prompting techniques and LLM agent works from the literature. Your goal is to maximize "fitness" by proposing interestingly new agents. 
Observe the discovered architectures carefully and think about what insights, lessons, or stepping stones can be learned from them.
Please focus on the architecture with the optimal fitness, and based on that, propose what you believe is the most likely next agent architecture. Note that each optimization step can involve adding one or two new modules to the current best solution, or proposing an entirely novel solution. However, it's important to ensure that each change remains relatively simple and not overly complex.
\end{lstlisting}

\subsubsection{Outer Loop Workflow Optimization Prompt}
\label{subsubsec:outer_prompt}

\begin{lstlisting}[style=promptstyle]
# Overview
You are an expert machine learning researcher testing various agentic systems. Your objective is to design building blocks such as prompts and control flows within these systems to solve complex tasks. Your aim is to design an optimal agent performing well on the MATH dataset, which evaluates mathematical problem-solving abilities across various mathematical domains including algebra, counting and probability, geometry, intermediate algebra, number theory, prealgebra and precalculus.

## An example question from MATH:

**instruction (Not Given)**: Solve the following problem and provide a detailed solution. Present the final answer using the \\boxed{} format.

**question**: question

**solution (Not Given)**: solution 

Note: We divide the overall MATH task into seven distinct subtasks. Below is the performance of the Discovered Architecture Archive on each of these seven subtasks.
Discovered Architecture Archive
The following presents the archive of the discovered architectures on seven subtasks as well as the full MATH task:

[ARCHIVE_LIST]

The fitness value is defined as the accuracy on a validation question set. Your goal is to identify an architecture that either maximizes fitness across the seven subtasks or can quickly evolve toward that goal. Note that you should not limit yourself to only the most recently generated architectures—your objective is to maximize this fitness.
# Output Instruction and Example:
The first key should be ("thought"), and it should capture your thought process for designing the next function. In the "thought" section, first reason about what should be the next interesting agent to try, then describe your reasoning and the overall concept behind the agent design, and finally detail the implementation steps.
The second key ("name") corresponds to the name of your next agent architecture. 
Finally, the last key ("code") corresponds to the exact “forward()” function in Python code that you would like to try. You must write a COMPLETE CODE in "code": Your code will be part of the entire project, so please implement complete, reliable, reusable code snippets.

Here is an example of the output format for the next agent architecture:

[EXAMPLE]

You must use the exact function interface used above. You need to specify the instruction, input information, and the required output fields for various LLM agents to do their specific part of the architecture. 
Also, it could be helpful to set the LLM’s role and temperature to further control the LLM’s response. Note that the LLMAgentBase() will automatically parse the output and return a list of “Infos”. You can get the content by Infos.content. 
DO NOT FORGET the taskInfo input to LLM if you think it is needed, otherwise LLM will not know about the task.

## WRONG Implementation examples:
Here are some mistakes you may make:

1. This is WRONG: ```
feedback, correct = critic_agent([taskInfo, thinking, answer], critic_instruction, i)
feedback_info = verifier_agent([taskInfo, Info('feedback', 'Critic Agent', thinking, 0)], verification_instruction)
```
It is wrong to use "Info('feedback', 'Critic Agent', thinking, 0)". The returned "feedback" from LLMAgentBase is already Info.

# Your task
You are well-versed in LLM prompting techniques and agent-based frameworks from the literature. You are tasked with designing a new agent architecture based on the best-performing solutions from each subtask of the MATH benchmark. The goal is for this new architecture to satisfy at least one of the following criteria:

It effectively integrates key modules and features from the optimal solutions of individual subtasks, resulting in a generalizable and adaptable architecture that performs well across all subtasks;

Alternatively, the architecture should exhibit strong adaptability and rapid update capabilities, allowing it to quickly evolve and converge toward the optimal solution for each specific subtask.
However, you should ensure that the newly generated frameworks is not significantly more complex than the original one, and you may also remove some redundant LLM invocation code.
\end{lstlisting}

\subsubsection{Reflection Prompt}
\begin{lstlisting}[style=promptstyle]
We noticed that the current agent is prone to making mistakes when handling the following cases:
[CASE_LIST]

Please analyze the reasons for these mistakes and propose improvements.

Your response should be organized as follows:

"reflection": Provide your thoughts on the mistakes in the implementation, and suggest improvements.

"thought": Revise your previous proposal or propose a new architecture if necessary, using the same format as the example response.

"name": Provide a name for the revised or new architecture. (Don't put words like "new" or "improved" in the name.)

"code": Provide the corrected code or an improved implementation. Make sure you actually implement your fix and improvement in this code.
\end{lstlisting}

\subsubsection{Test-Time Adaptation Workflow Optimization Prompt}
\label{subsubsec:adapt_prompt}
\begin{lstlisting}[style=promptstyle]
# Overview
You are an expert machine learning researcher testing various agentic systems. Your objective is to design building blocks such as prompts and control flows within these systems to solve complex tasks. Your goal is to design an optimal agent that performs well on the MATH dataset. You may analyze the characteristics of these problems and then design an agent capable of effectively solving them.

[TASK_DSC]

Note: Your goal is to design an improved agent based on the previous agent, tailored to the characteristics of the current task. We aim to rapidly enhance the performance of the current agent.

# Output Instruction and Example:
The first key should be ("thought"), and it should capture your thought process for designing the next function. In the "thought" section, first reason about what should be the next interesting agent to try, then describe your reasoning and the overall concept behind the agent design, and finally detail the implementation steps.
The second key ("name") corresponds to the name of your next agent architecture. 
Finally, the last key ("code") corresponds to the exact “forward()” function in Python code that you would like to try. You must write a COMPLETE CODE in "code": Your code will be part of the entire project, so please implement complete, reliable, reusable code snippets.

Here is an example of the output format for the next agent architecture:

[EXAMPLE]

You must use the exact function interface used above. You need to specify the instruction, input information, and the required output fields for various LLM agents to do their specific part of the architecture. 
Also, it could be helpful to set the LLM’s role and temperature to further control the LLM’s response. Note that the LLMAgentBase() will automatically parse the output and return a list of “Infos”. You can get the content by Infos.content. 
DO NOT FORGET the taskInfo input to LLM if you think it is needed, otherwise LLM will not know about the task.

# Your task
You are well-versed in LLM prompting techniques and agent-based frameworks from the literature. You are tasked with designing a new agent architecture based on the previous agent to solve the current task.
\end{lstlisting}

\subsection{Workflow Case}
To provide a concrete illustration of our system's output, we present the workflow code generated in the final outer-loop iteration on the \textsc{MATH} dataset. This example reflects the culmination of iterative refinement across subtasks and highlights the integration of shared and task-specific modules.

\begin{lstlisting}[style=promptstyle]
def forward(self, taskInfo):
    import re
    from collections import Counter

    def extract(text):
        for p in [r'\\boxed{([^}]*)}', r'\(([^)]+)\)', r'\\frac{[^}]*}{[^}]*}', r'(\d+)\s*$']:
            m = re.search(p, text)
            if m: return m.group(0).strip()

    roles = ['Math Professor', 'Grade School Teacher', 'Math Enthusiast', 'Math Olympiad Student', 'Helpful Assistant']
    agents = [LLMAgentBase(['thinking', 'solution'], f'A{i}', role=r, temperature=0.7 + 0.1*i) for i, r in enumerate(roles)]
    sols = [a([taskInfo], "Please think step by step and solve.", i) for i, a in enumerate(agents)]

    ext_agent = LLMAgentBase(['extracted_answer'], 'Extractor', role='Answer Extractor', temperature=0.1)
    answers, amap = [], {}
    for i, (t, s) in enumerate(sols):
        ans = ext_agent([taskInfo, s], "Extract ONLY final boxed answer.", i)[0].content.strip() or extract(s.content)
        if ans: answers.append(ans); amap.setdefault(ans, (t, s))

    top = Counter(answers).most_common()
    if top:
        top_answers = [a for a, c in top if c == top[0][1]]
        if len(top_answers) == 1:
            _, sol = amap[top_answers[0]]
        else:
            inputs = [taskInfo] + sum((list(amap[a]) for a in top_answers), []) + [Info('extracted_answer', '', a, -1) for a in top_answers]
            sol = LLMAgentBase(['thinking', 'solution'], 'Final Decider', temperature=0.1)(inputs, "Choose best answer.")[1]
    else:
        inputs = [taskInfo] + sum(([t, s] for t, s in sols), [])
        sol = LLMAgentBase(['thinking', 'solution'], 'Fallback Decider', temperature=0.1)(inputs, "Choose among all.")[1]

    verifier = LLMAgentBase(['feedback', 'correct'], 'Verifier', role='Checker', temperature=0.1)
    clarifier = LLMAgentBase(['clarification'], 'Clarifier', role='Solver', temperature=0.4)
    synthesizer = LLMAgentBase(['thinking', 'solution'], 'Synth', temperature=0.3)

    for i in range(2):
        ext = ext_agent([taskInfo, sol], "Extract ONLY final boxed answer.", 100+i)[0]
        fb, ok = verifier([taskInfo, sol, ext], "Check correctness.", i)
        if ok.content == 'True': return sol

        clar, = clarifier([taskInfo, sol, fb], "Respond to critique.", i)
        fb2, ok2 = verifier([taskInfo, sol, ext, clar], "Recheck solution.", 100+i)
        if ok2.content == 'True': return sol

        syn_inputs = [taskInfo, sol, fb2, clar] + sum(sols, []) + [Info('extracted_answer', '', a, -1) for a in answers if a]
        sol = synthesizer(syn_inputs, "Revise or synthesize.")[1]

    return sol
\end{lstlisting}

\end{document}